\documentclass[draftcls, doublecolumn]{IEEEtran}

\date{}

\usepackage{blindtext}
\usepackage{graphicx}
\usepackage{array}
\usepackage[numbers,sort&compress]{natbib}
\usepackage{amsmath}
\usepackage{color}
\usepackage{algorithm2e}
\usepackage{subcaption}
\let\labelindent\relax
\usepackage{enumitem}
\usepackage{url}
\usepackage{multirow}
\usepackage{multicol}
\usepackage{hyperref} 
\usepackage{xcolor}
\usepackage{adjustbox}
\usepackage[geometry]{ifsym}
\usepackage{hyperref}
\usepackage{textcomp}
\definecolor{blue_1}{RGB}{204, 229, 255}

\definecolor{blue_2}{RGB}{220, 240, 255}

\begin{document}

\onecolumn
\begin{itemize}[align=parleft, labelsep=2.0cm]

\item[\textbf{Citation}]{D. Temel and G. AlRegib, "Traffic Signs in the Wild: Highlights from the IEEE Video and Image Processing Cup 2017 Student Competition [SP Competitions]," in IEEE Signal Processing Magazine, vol. 35, no. 2, pp. 154-161, March 2018.}

\item[\textbf{DOI}]{https://doi.org/10.1109/MSP.2017.2783449}

\item[\textbf{Review}]{Date of publication: 7 March 2018}

\item[\textbf{VIP}]{https://ghassanalregib.com/vip-cup/}

\item[\textbf{Bib}] {@ARTICLE\{Temel2018\_SPM,\\ 
author=\{D. Temel and G. AlRegib\},\\ 
journal=\{IEEE Signal Processing Magazine\},\\ 
title=\{Traffic Signs in the Wild: Highlights from the IEEE Video and Image Processing Cup 2017 Student Competition [SP Competitions]\}, \\ 
year=\{2018\},\\ 
volume=\{35\},\\ 
number=\{2\},\\ 
pages=\{154-161\},\\ 
keywords=\{traffic engineering computing;video signal processing;IEEE Video and Image Processing Cup 2017 Student Competition;traffic signs\},\\ 
doi=\{10.1109/MSP.2017.2783449\},\\ 
ISSN=\{1053-5888\},\\ 
month=\{March\},\}
} 

\item[\textbf{Copyright}]{\textcopyright 2018 IEEE. Personal use of this material is permitted. Permission from IEEE must be obtained for all other uses, in any current or future media, including reprinting/republishing this material for advertising or promotional purposes,
creating new collective works, for resale or redistribution to servers or lists, or reuse of any copyrighted component
of this work in other works. }

\item[\textbf{Contact}]{alregib@gatech.edu~~~~~~~\url{https://ghassanalregib.com/}\\dcantemel@gmail.com~~~~~~~\url{http://cantemel.com/}}
\end{itemize}
\thispagestyle{empty}
\newpage
\clearpage
\setcounter{page}{1}

\twocolumn

\title{\huge  Traffic Signs in the Wild: Highlights from the IEEE Video and Image Processing Cup 2017 Student Competition [SP Competitions]}

\author{Dogancan Temel and Ghassan AlRegib}

\maketitle

\IEEEpeerreviewmaketitle

As we witness the fourth industrial revolution, several aspects of our daily lives will soon be impacted beyond recognition. The list includes healthcare, education, security, transportation, warfare, and entertainment. Transportation, in particular, is undergoing a set of disruptive technologies including electrical vehicles (EV) and autonomous vehicles (AV). Although AVs have witnessed a revolution in many aspects over the past twenty years, deploying AVs in the wild remains to be a challenge. One of the basic features of AVs is to understand the surroundings and interpret sensed data. This requires the deployment of recognition algorithms that are expected to operate under all conditions. One of the most researched recognition applications in the literature is traffic sign recognition (TSR). Nevertheless, testing TSR algorithms under challenging conditions has been lagging for a number of reasons. One major factor is the limitation of existing datasets in terms of challenging conditions and metadata. To address such shortcomings, the \textbf{C}hallenging \textbf{U}nreal and \textbf{R}eal \textbf{E}nvironments for \textbf{T}raffic \textbf{S}ign \textbf{D}etection (\texttt{CURE-TSD}) dataset  was recently  introduced~\cite{curetsd_dataset}, which was also utilized for traffic sign recognition in \cite{Temel2017_NIPSW}. 

The \texttt{CURE-TSD} dataset was used to host the first edition of the Video and Image Processing (VIP) Cup in 2017 denoted as \emph{Traffic Sign Detection under Challenging Conditions}.
The VIP Cup is a student competition in which undergraduate students form teams to work on real-life challenges. Each team should include one faculty member as an advisor, at most one graduate student as a mentor, and at least three but no more than ten undergraduate students. Formed teams participate in an open competition and the top three teams are selected to present their work at the final competition, which was held at the 2017 \textit{IEEE International Conference on Image Processing} (ICIP) in Beijing, China. Travel costs of finalist teams were supported by the IEEE Signal Processing Society (SPS). 

In this article, we share an overview of the VIP Cup experience including competition setup, teams, technical approaches, statistics, and competition experience through finalist teams members' and organizers' eyes.   

\begin{figure*}[htbp!]
\begin{adjustbox}{minipage=\linewidth-4pt,margin=5pt 5pt,bgcolor=blue_2,frame=0pt}
\centering
\begin{minipage}[b]{0.32\linewidth}
  \centering
\includegraphics[trim=0cm 1.5cm 0cm 0cm,width=\linewidth]{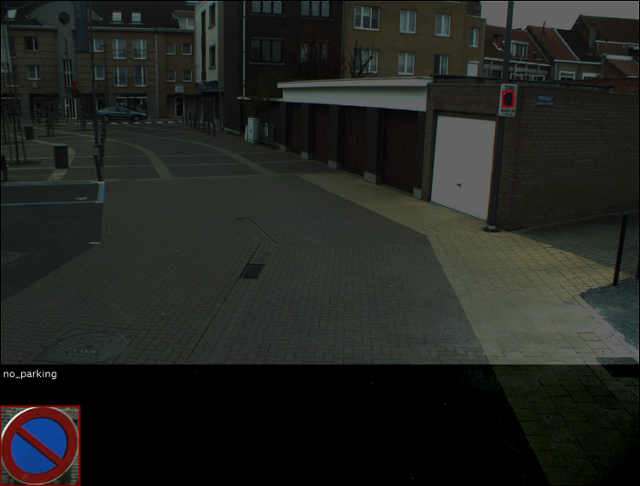}
  \vspace{0.15cm}
  \centerline{\footnotesize{(a) Darkness }}
\end{minipage}
\begin{minipage}[b]{0.32\linewidth}
  \centering
\includegraphics[trim=0cm 1.5cm 0cm 0cm,width=\linewidth]{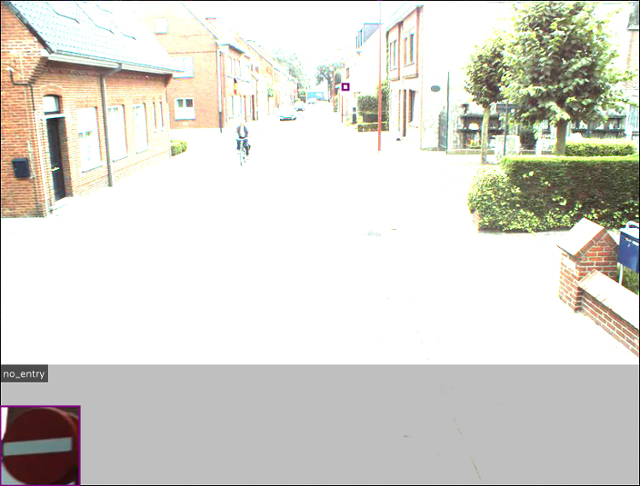}
  \vspace{0.15cm}
  \centerline{\footnotesize{(b) Brightness }}
\end{minipage}
\begin{minipage}[b]{0.32\linewidth}
  \centering
\includegraphics[trim=0cm 1.5cm 0cm 0cm,width=\linewidth]{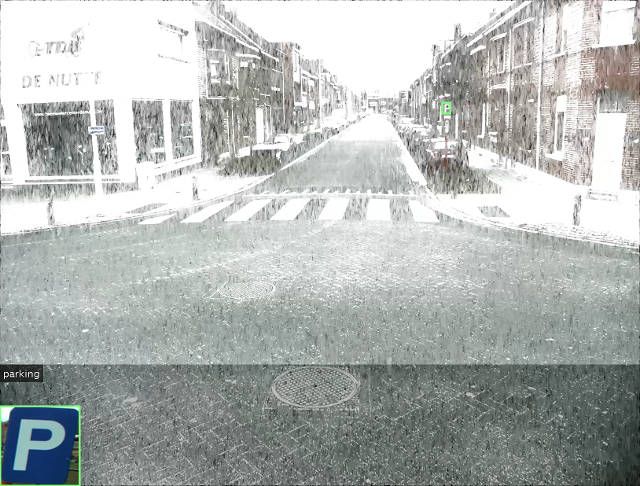}
  \vspace{0.15cm}
  \centerline{\footnotesize{(c) Snow }}
\end{minipage}
\begin{minipage}[b]{0.32\linewidth}
  \centering
\includegraphics[trim=0cm 1.5cm 0cm 0cm,width=\linewidth]{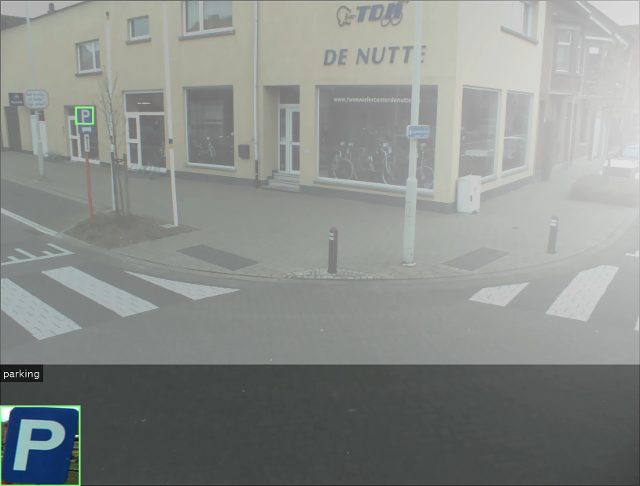}
  \vspace{0.15cm}
  \centerline{\footnotesize{(d) Haze }}
\end{minipage}
\begin{minipage}[b]{0.32\linewidth}
  \centering
\includegraphics[trim=0cm 1.5cm 0cm 0cm,width=\linewidth]{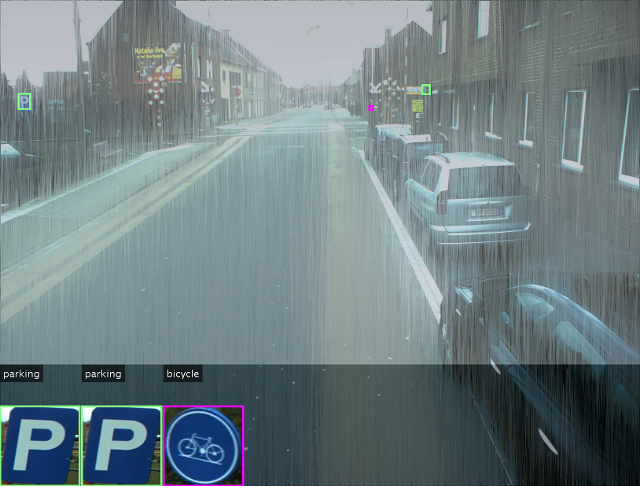}
  \vspace{0.15cm}
  \centerline{\footnotesize{(e) Rain }}
\end{minipage}
\begin{minipage}[b]{0.32\linewidth}
  \centering
\includegraphics[trim=0cm 1.5cm 0cm 0cm,width=\linewidth]{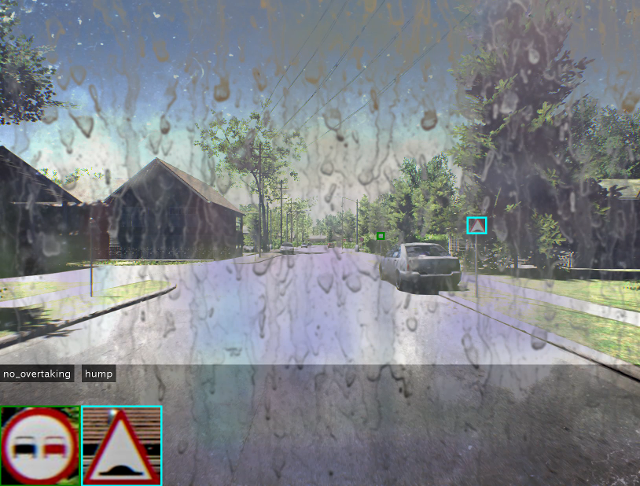}
  \vspace{0.15cm}
  \centerline{\footnotesize{(f) Dirty lens }}
\end{minipage}
\begin{minipage}[b]{0.32\linewidth}
  \centering
\includegraphics[trim=0cm 1.5cm 0cm 0cm,width=\linewidth]{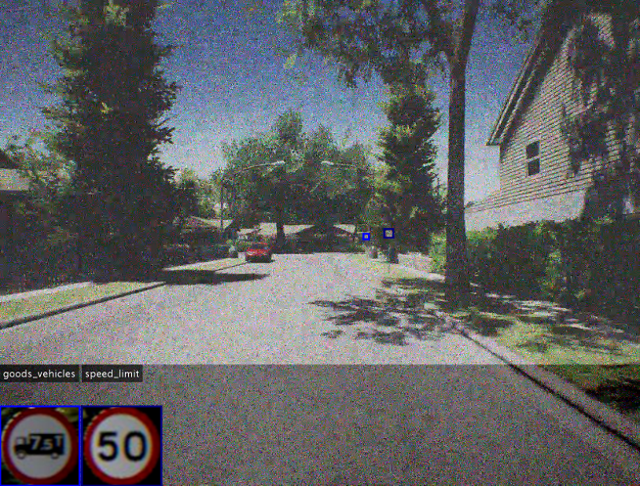}
  \vspace{0.15cm}
  \centerline{\footnotesize{(g) Noise }}
\end{minipage}
\begin{minipage}[b]{0.32\linewidth}
  \centering
\includegraphics[trim=0cm 1.5cm 0cm 0cm,width=\linewidth]{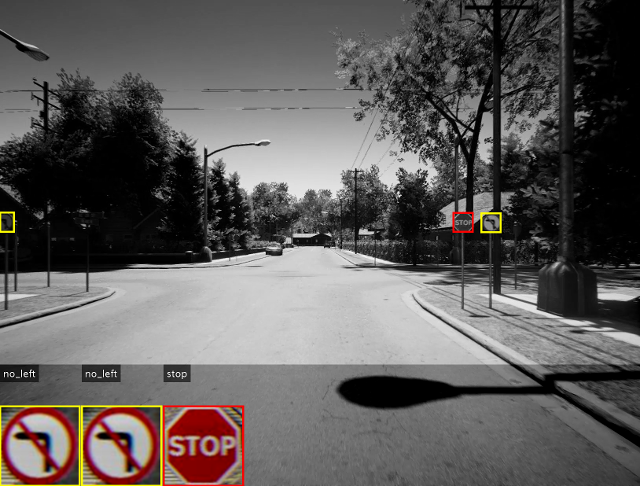}
  \vspace{0.15cm}
  \centerline{\footnotesize{(h) Decolorization }}
\end{minipage}
\begin{minipage}[b]{0.32\linewidth}
  \centering
\includegraphics[trim=0cm 1.5cm 0cm 0cm,width=\linewidth]{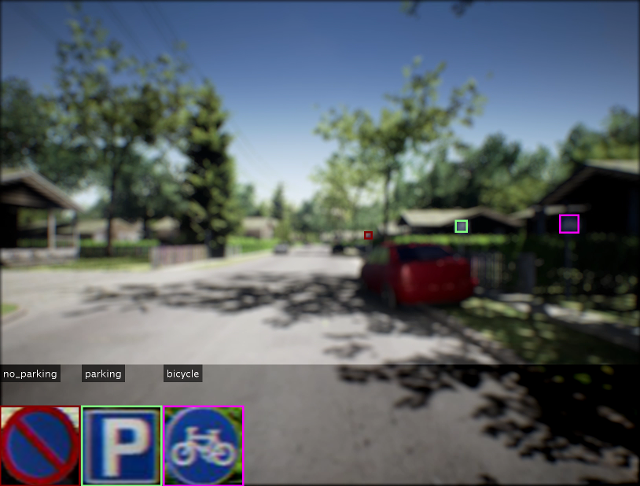}
  \vspace{0.15cm}
  \centerline{\footnotesize{(i) Blur }}
\end{minipage}
\vspace{-0.2cm}
\caption{Challenging scene examples from the 2017 VIP cup.}
\label{fig:scenes}
\vspace{-0.5cm}
\end{adjustbox}
\end{figure*}

\vspace{5.0mm}

\noindent \textbf{Traffic Sign Recognition under Challenging Conditions:}
Traffic signs can be recognized by state-of-the-art algorithms with high precision and accuracy in existing datasets, which are limited in terms of challenging conditions and corresponding metadata. The limited nature of these test sets makes it difficult to estimate the performance of recognition algorithms in non-ideal real-world scenarios. Recent studies \cite{Lu2017,Das2017} showed that adversarial perturbations can degrade the performance of existing traffic sign recognition systems under specific conditions. Even though these studies shed a light on conditions that are intentionally designed to fool existing systems, introduced non-idealities are inherently different from realistic challenging conditions. To perform practical robustness tests for traffic sign detection and recognition systems, we need to test them with realistic mild-to-severe challenges, which is the main objective of the VIP Cup 2017. 
The challenges in the VIP Cup include multiple levels of rain, snow, haze, brightness, darkness, shadow, blur, decolorization, codec error, dirty lens, and noise, whose examples are depicted in  Fig.~\ref{fig:scenes}.

\vspace{5.0mm}

\begin{figure*}[htbp!]
\centering
\begin{minipage}[b]{\linewidth}
  \centering
\includegraphics[width=\linewidth, trim= 10mm 20mm 10mm 5mm]{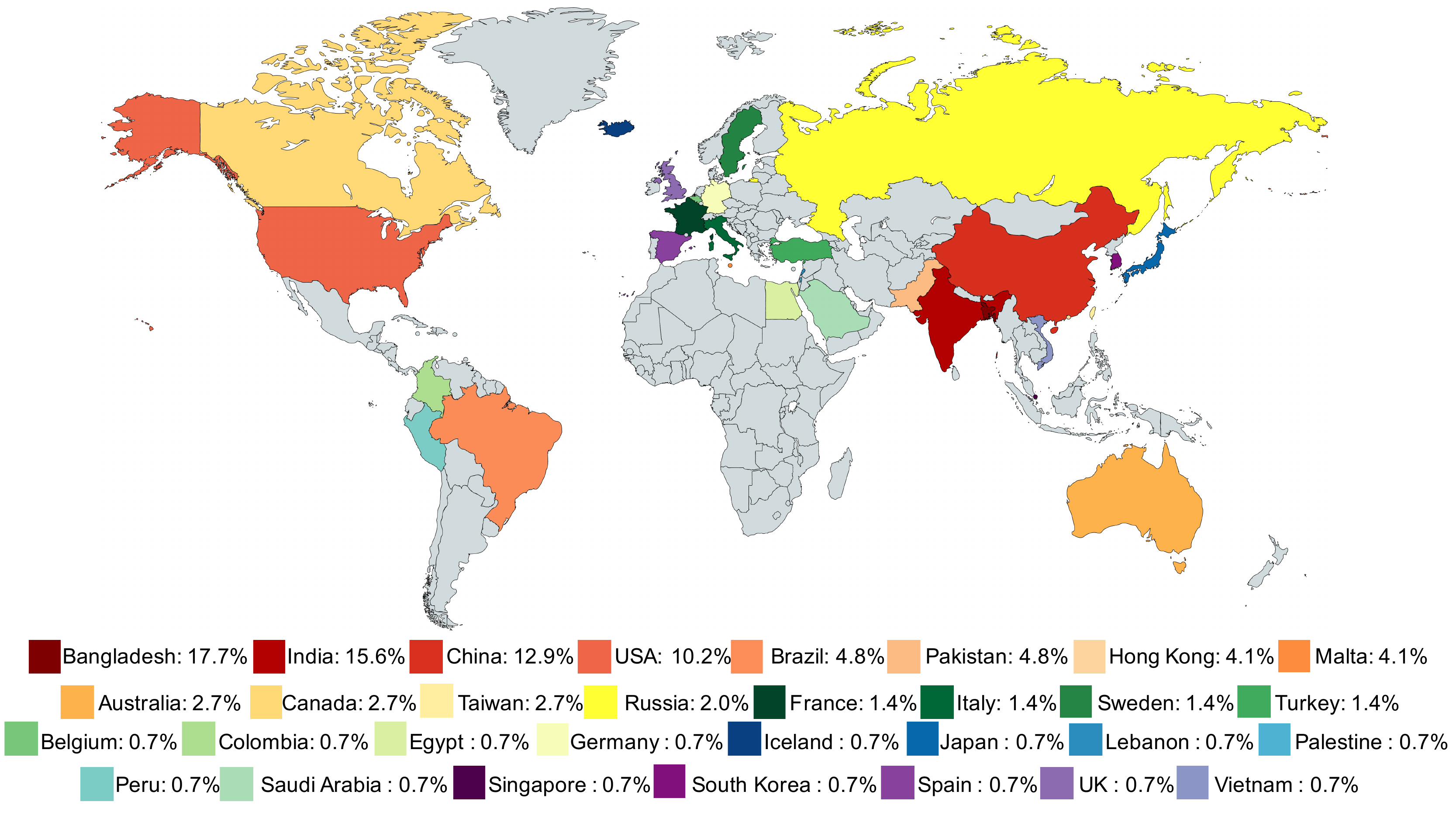}
  \vspace{0.01cm}
\end{minipage}
\caption{Global engagement map of the VIP Cup 2017.}
\label{fig:world_map}
\end{figure*}

\noindent \textbf{VIP Cup 2017 Statistics:} The VIP Cup 2017 started with a global engagement of more than 250 requests from 147 parties to access competition data from all around the world as shown in Fig.~\ref{fig:world_map}. At the start line, the highest engagement was received from Bangladesh, India, China, and USA. At the registration stage, there were 80 members clustered into 19 teams from 10 countries including Australia, Bangladesh (2 teams), China (7 teams), Hong Kong (2 teams), India, Malta, Pakistan, Sweden, Taiwan, and USA (2 teams). Out of these 19 teams, 6 teams with a total of 32 members  from Australia, Bangladesh (2 teams), China, Hong Kong, Sweeden, and Taiwan made it to the final stage. 

\vspace{5.0mm}

\noindent \textbf{Tasks in the VIP Cup 2017:} The VIP Cup 2017 included an open competition stage and a final stage. The call for competition was announced on February 15, 2017 and the open competition started by making data publicly available on March 15, 2017. The video dataset was released in the open competition stage, which included processed versions of captured and synthesized traffic videos with challenging conditions spanning a wide range from mild to severe. The competition dataset was split into $70\%$ training set and $30\%$ test set.  Specifically, $3,978$ sequences were provided for model development and $1,755$ sequences for final testing. There were $300$ frames in each video sequence. Traffic signs within the video sequences included \emph{bicycle, goods vehicles, hump, no entry, no left, no overtaking, no parking, no right, no stopping, parking, priority to, speed limit, stop, and yield}. The participants were asked to develop algorithms that can detect these traffic signs under challenging conditions in the test set, which cannot be used in the model development. Participants were allowed to use Maltab\textsuperscript{\textregistered} as a coding platform and Python and C++ as coding languages along with any library or toolboxes. Competition rules, which were set to have a fair competition ground, guarantee reproducible research, and obtain practical algorithms, are as follows:         
\begin{itemize}
    \item Any algorithm that utilizes future frames for prediction will be disqualified.
    \item Any algorithm that utilizes testing labels in the final evaluation will be disqualified.
    \item Any algorithm that utilizes testing sequences or labels in the training including model training and validation will be disqualified.
    \item The submissions should include detailed instructions and necessary codes to replicate the results. Otherwise, the participants can be disqualified.
    \item Reproducing results including training, testing, or any other processes should not exceed a reasonable amount of time that allows the organizers to evaluate all submissions within the given time window.
\end{itemize}

The open competition stage was completed on July 8, 2017, which was the deadline to receive team submissions that included:
\begin{itemize}
    \item a report in the form of an IEEE conference paper up to six pages, on the technical details of the methods used, programs developed, and results;
    
    \item estimated detection files for each test sequence; and
    
    \item all codes with detailed comments and \texttt{README} files. 
    
\end{itemize}

The VIP Cup 2017 organizers evaluated the submissions and announced the finalists as team \texttt{Neurons}, team \texttt{PolyUTS}, and team \texttt{Markovians} on August 15, 2017. Evaluation was based on overall precision, recall, and combination of these metrics. Finalist teams were invited to the final competition at the 2017 \textit{IEEE International Conference on Image Processing}, which was held in Beijing, China, September 17-20, 2017. Finalist teams presented their work on Sunday, September 17, 2017. Each team had 15 minutes for their presentation and 5 minutes for questions and answers. After team presentations, the jury had an internal discussion to finalize the ranking. In the opening ceremony of the conference on September 18, 2017, IEEE SPS president Dr. Rabab Ward highlighted the first Video and Image Processing Cup and publicly announced the winners of the competition. The jury included Dr. Amy Reibman, Dr. Béatrice Pesquet, Dr. Patrizio Campisi, and Dr. Ghassan AlRegib.



\begin{figure*}[htbp!]
\begin{adjustbox}{minipage=\linewidth-4pt,margin=5pt 30pt,bgcolor=blue_1,frame=0pt}
\vspace{-0.8cm}
\centering
\begin{minipage}[b]{0.48\linewidth}
  \centering
\includegraphics[trim=0cm 0.25cm 0cm 0cm,width=\linewidth]{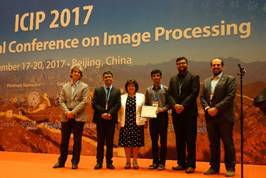}
  \vspace{0.15cm}
\centerline{\footnotesize{(a) }}
\end{minipage}
\begin{minipage}[b]{0.48\linewidth}
  \centering
\includegraphics[trim=0cm 1.2cm 0cm 0cm,width=\linewidth]{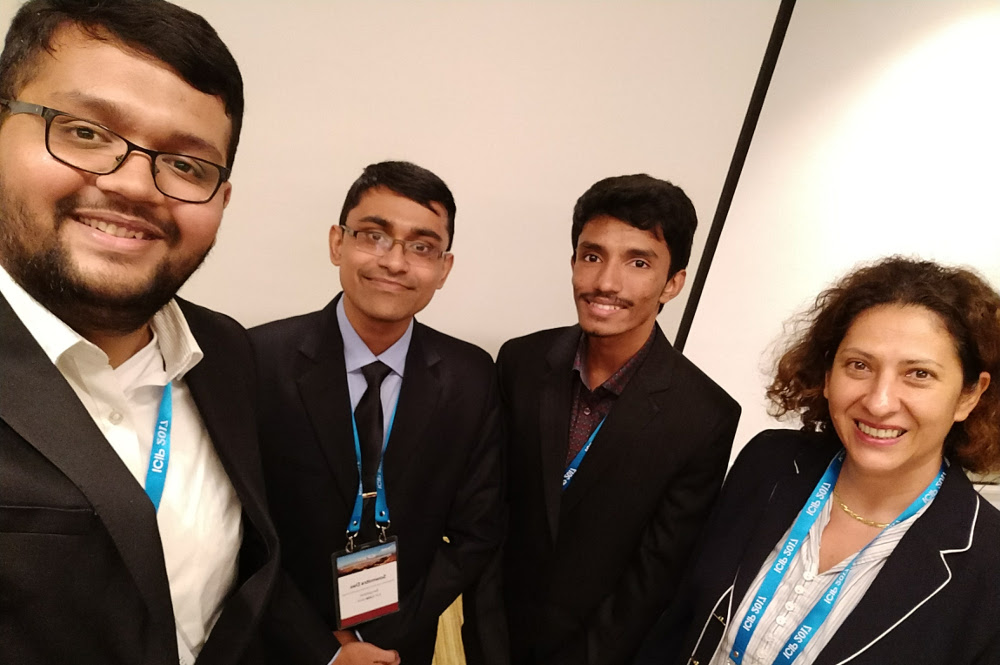}
  \vspace{0.15cm}
\centerline{\footnotesize{(b) }}
\end{minipage}
\begin{minipage}[b]{0.48\linewidth}
  \centering
\includegraphics[trim=0cm 1.cm 0cm 0cm,width=\linewidth]{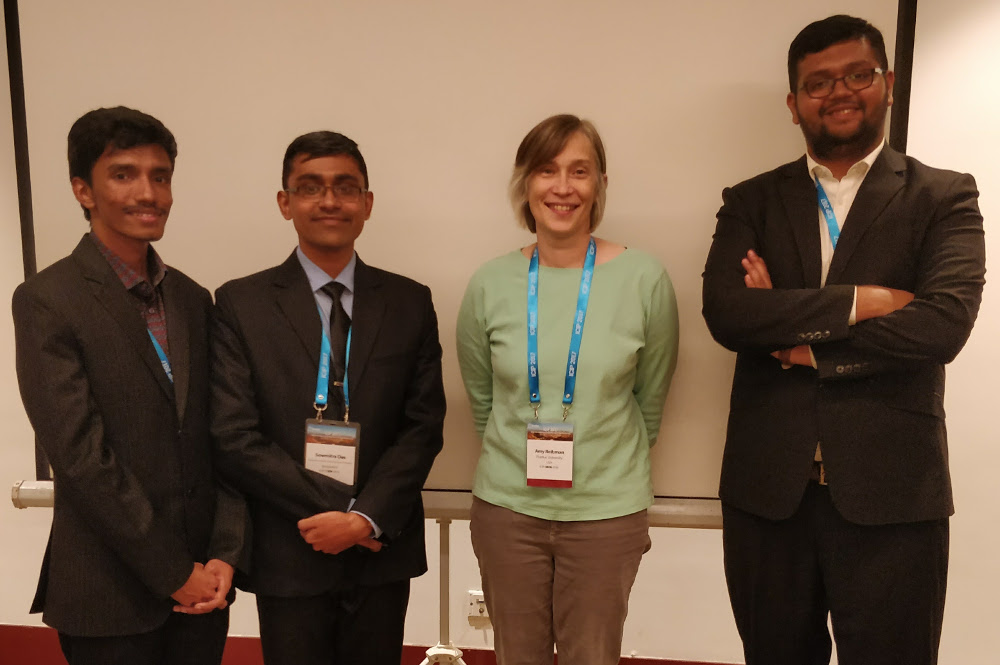}
  \vspace{0.15cm}
\centerline{\footnotesize{(c) }}
\end{minipage}
\begin{minipage}[b]{0.48\linewidth}
  \centering
\includegraphics[trim=0cm 1.cm 0cm 0cm,width=\linewidth]{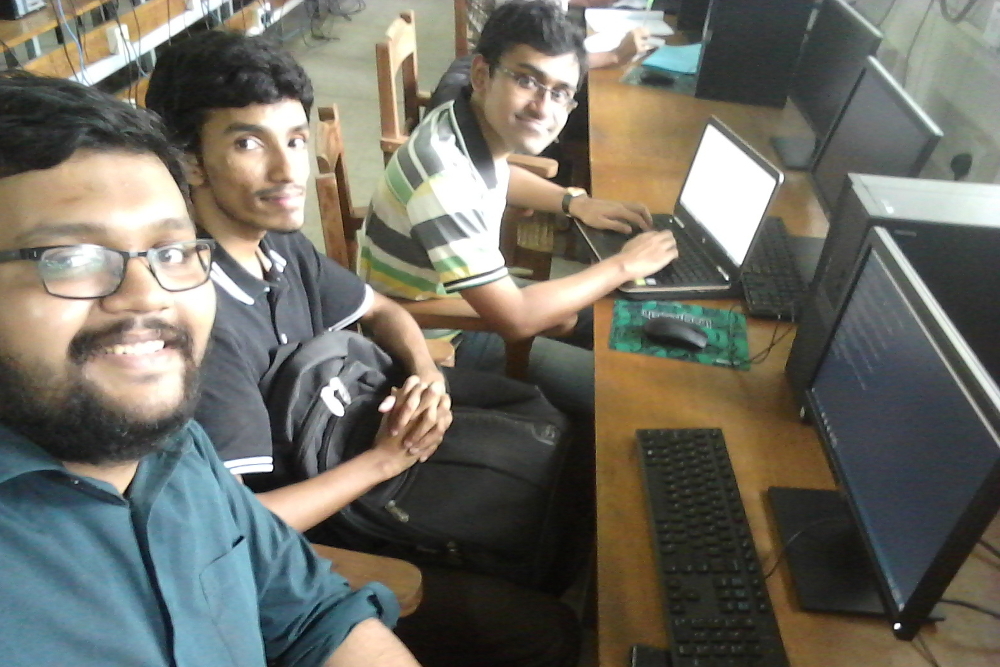}
  \vspace{0.15cm}
\centerline{\footnotesize{(d) }}
\end{minipage}

\caption{First team: \texttt{Neurons} (a) with IEEE SPS president Dr. Rabab Ward, IEEE student services director Dr. Patrizio Campisi and VIP Cup 2017 organizer Dr. Ghassan AlRegib, (b) with judge Dr. B\'eatrice Pesquet-Popescu, (c) with judge Dr. Amy R. Reibman, and (d) behind the scenes. }
 \caption*{\parbox[b][10em][t]{0.85\textwidth}{\textbf{Affiliation:} Bangladesh University of Engineering and Technology \\
     \textbf{Undergraduate students:} Uday Kamal, Sowmitra Das, Abid Abrar \\
    \textbf{Supervisor:} Md Kamrul Hassan \\
    \textbf{Technical Approach:} Team \texttt{Neurons} developed a data-driven system based on Convolutional Neural Networks (CNNs) similar to an existing approach \cite{Simonyan2015} to identify the type of challenging conditions in a scene. Based on the identified challenging condition, they performed a preprocessing operation over video frames to eliminate the effect of challenging conditions and enhance traffic sign visibility. After the preprocessing stage, they trained sepate CNNs to localize and recognize traffic signs. In their algorithm development, they used the Keras API with Tensoflow back-end on NVIDIA GeForce GTX 1050 GPU.  }
    }

\label{fig:winners_neurons}
\vspace{0.5cm}
\end{adjustbox}

\end{figure*}


\begin{figure*}[htbp!]
\begin{adjustbox}{minipage=\linewidth-4pt,margin=5pt 30pt,bgcolor=blue_1,frame=0pt}
\vspace{-0.8cm}
\centering
\begin{minipage}[b]{0.48\linewidth}
  \centering
\includegraphics[trim=0cm 0.25cm 0cm 0cm,width=\linewidth]{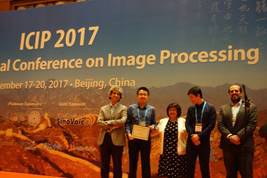}
  \vspace{0.15cm}
 \centerline{\footnotesize{(a) }}
\end{minipage}
\begin{minipage}[b]{0.48\linewidth}
  \centering
\includegraphics[trim=0cm 0.85cm 0cm 0cm,width=\linewidth]{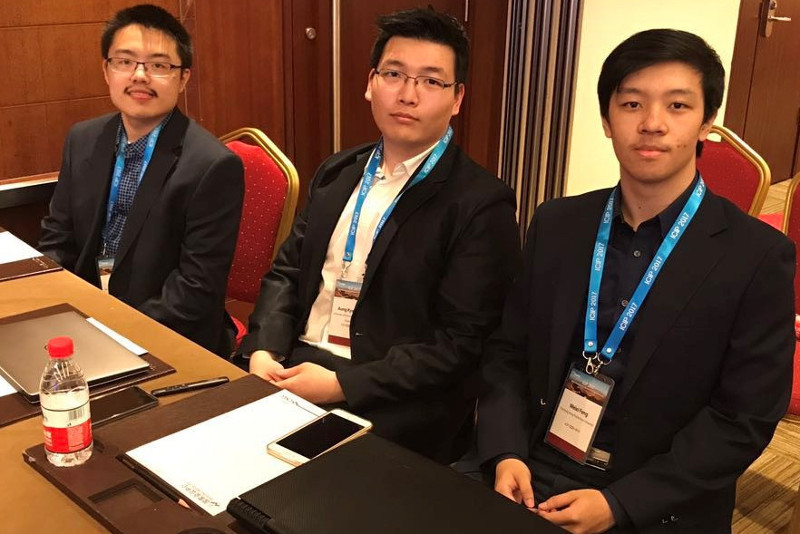}
  \vspace{0.15cm}
 \centerline{\footnotesize{(b) }}
\end{minipage}
\begin{minipage}[b]{0.48\linewidth}
  \centering
\includegraphics[trim=0cm 1.0cm 0cm 0cm,width=\linewidth]{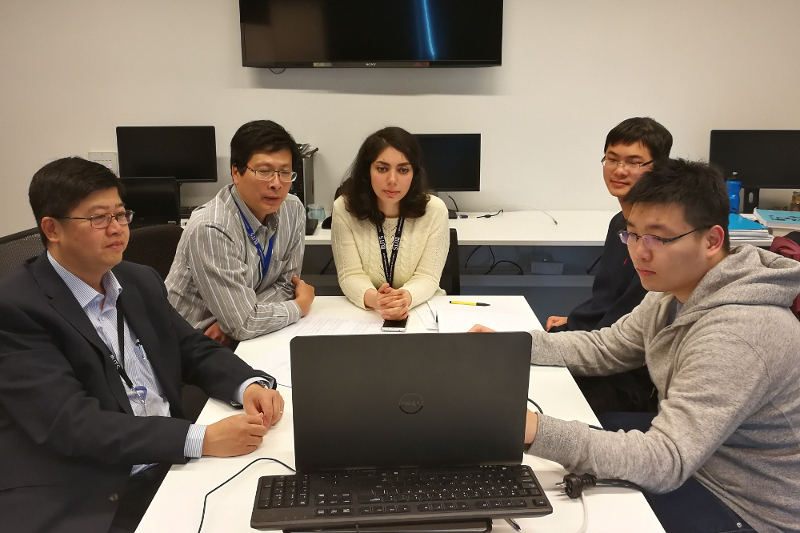}
  \vspace{0.15cm}
 \centerline{\footnotesize{(c) }}
\end{minipage}
\begin{minipage}[b]{0.48\linewidth}
  \centering
\includegraphics[trim=0cm 1.0cm 0cm 0cm,width=\linewidth]{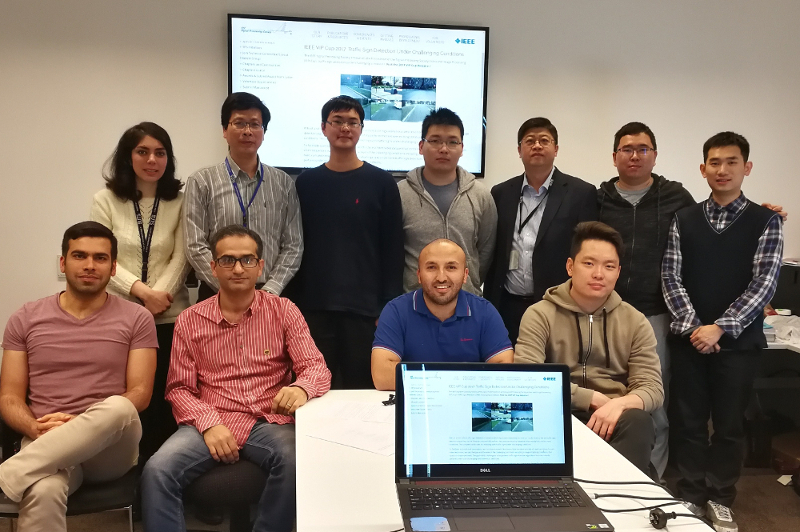}
  \vspace{0.15cm}
 \centerline{\footnotesize{(d) }}
\end{minipage}

\caption{First runner-up: \texttt{PolyUTS} (a) with IEEE SPS president Dr. Rabab Ward, IEEE student services director Dr. Patrizio Campisi and VIP Cup 2017 organizer Dr. Ghassan AlRegib,  (b) VIP Cup 2017 final, (c-d) behind-the-scenes. }
\caption*{    \parbox[b][10em][t]{0.85\textwidth}{\textbf{Affiliation:} University of Technology Sydney, Hong Kong Polytechnic University, University of New South Wales \\
    \textbf{Undergraduate students:} Weixi Feng, Aung Min, Jiawei Zhang, Chenhang He, Hardy Zhu, Wenqi Jia \\
    \textbf{Supervisor:} Xiangjian He \\
    \textbf{Technical Approach:} Team \texttt{PolyUTS} utilized two Convolutional Neural Networks (CNN) trained with the competition dataset, one is for possible region proposal, and the other is for classification. Region proposal network was based on extracting image features with a pretrained CNN \cite{Szegedy2015} and regressing these features. To reduce the time consumed during the region proposal procedure, they forward propagated all possible bounding box coordinates once and used the second CNN architecture to classify proposed regions. In their algorithm development, they utilized Tensorflow and OpenCV on NVIDIA GTX 1080 GPU.}
    }
\label{fig:winners_polyuts}
\vspace{1.7cm}
\end{adjustbox}
\end{figure*}


\begin{figure*}[htbp!]
\begin{adjustbox}{minipage=\linewidth-4pt,margin=5pt 30pt,bgcolor=blue_1,frame=0pt}
\vspace{-0.8cm}
\centering
\begin{minipage}[b]{0.45\linewidth}
  \centering
\includegraphics[trim=0cm 0.25cm 0cm 0cm,width=\linewidth]{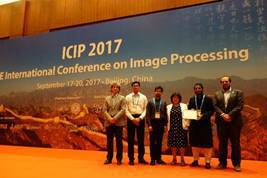}
  \vspace{0.15cm}
  \centerline{\footnotesize{(a) }}
\end{minipage}
\begin{minipage}[b]{0.45\linewidth}
  \centering
\includegraphics[trim=0cm 1.0cm 0cm 0cm,width=\linewidth]{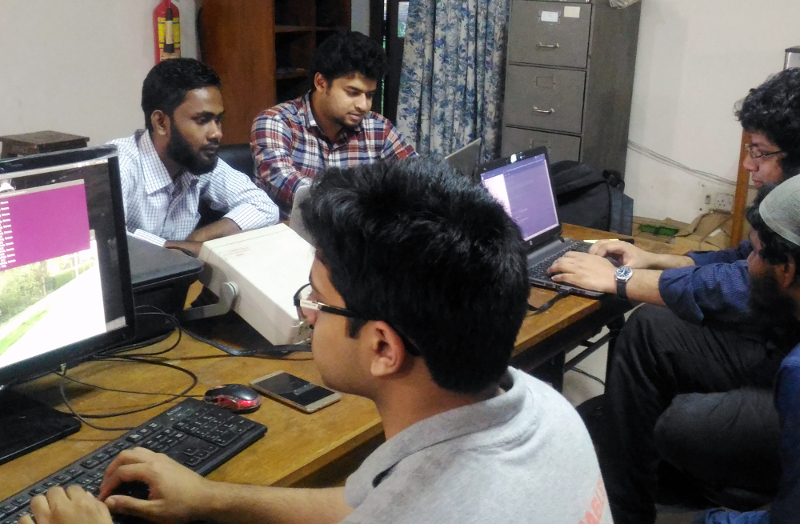}
  \vspace{0.15cm}
  \centerline{\footnotesize{(b) }}
\end{minipage}
\begin{minipage}[b]{0.45\linewidth}
  \centering
\includegraphics[trim=0cm 1.0cm 0cm 0cm,width=\linewidth]{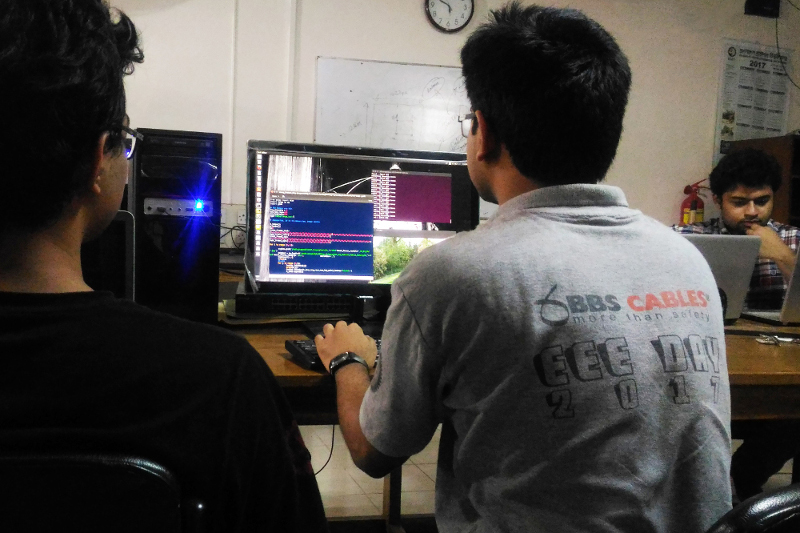}
  \vspace{0.15cm}
  \centerline{\footnotesize{(c) }}
\end{minipage}
\begin{minipage}[b]{0.45\linewidth}
  \centering
\includegraphics[trim=0cm 1.0cm 0cm 0cm,width=\linewidth]{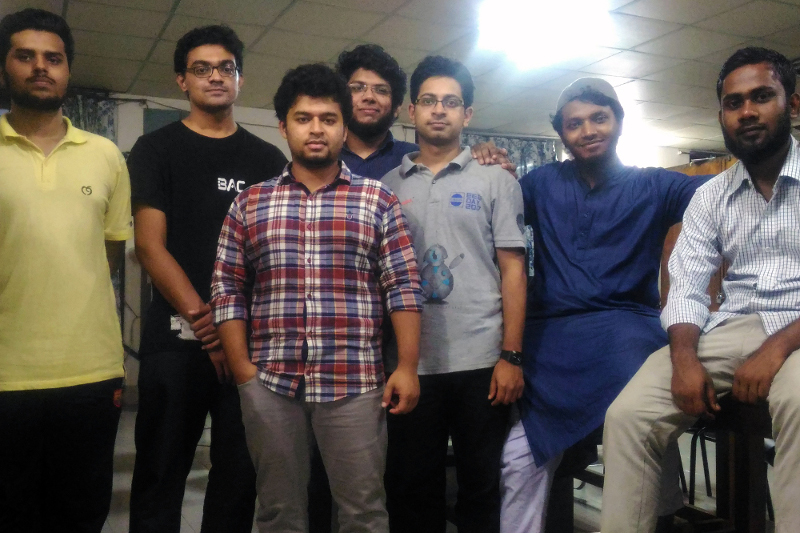}
  \vspace{0.15cm}
  \centerline{\footnotesize{(d) }}
\end{minipage}

\caption{Second runner-up: \texttt{Markovians} (a) with IEEE SPS president Dr. Rabab Ward, IEEE student services director Dr. Patrizio Campisi and VIP Cup 2017 organizer Dr. Ghassan AlRegib, (b-d) behind-the-scenes. }
\caption*{   \parbox[b][15em][t]{0.85\textwidth}{ \textbf{Affiliation:} Bangladesh University of Engineering and Technology \\
    \textbf{Undergraduate students:} Ahmed Maksud, Jubaer Hossain, Kinjol Barua, Roknuzzaman Rokon, Muhammad Suhail Najeeb, Nahian Ibn Hasan, Shahruk Hossain, Shakib Zaman, SM Raiyan Chowdhury \\
    \textbf{Graduate mentor:} Sayeed Shafayet Chowdhury \\ 
    \textbf{Supervisor:} Mohammad Ariful Haque \\
    \textbf{Technical Approach:} Team \texttt{Markovians} trained a recurrent Convolutional Neural Network (CNN) \cite{Liang2015} to identify challenging conditions in a scene and detected traffic signs with a faster region-based CNN architecture. They used a Kalman-based approach to track signs with static challenging conditions including dirty lens and shadow whereas they used a Lucas-Kanade-based approach for all other challenge types. They used a CNN architecture to recognize detected and tracked signs. In their algorithm development, they used Keras API with Tensorflow back-end and OpenCV.
    }
    }
\label{fig:winners_markovians}
\vspace{0.5cm}
\end{adjustbox}
\end{figure*}


\noindent \textbf{Highlights of technical approaches:}
All of the finalist algorithms are based on state-of-the-art data-driven methods. Specifically, baseline methods used by finalist algorithms rely on Convolutional Neural Networks (CNNs) that directly learn visual representations from examples with labels in a supervised fashion. In the VIP Cup, CNNs were utilized for various tasks including challenge type classification, preprocessing, localization, and recognition. The contribution of finalist algorithms to the literature can be considered as four folds. First, challenging conditions in video sequences were identified by team \texttt{Markovians} and  team \texttt{Nuerons}. Second, video sequences were processed with challenge-specific operations to enhance traffic sign visibility. Third, team \texttt{Markovians} tracked signs through temporal information, which is overlooked by majority of the state-of-the-art architectures in the literature. Forth, a challenge-aware tracking mechanism was used by team \texttt{Markovians}, which alternated tracking mechanism based on the dominant challenging condition in video sequences. The best performing method achieved a precision of $0.550$ and a recall of $0.320$ in the overall test set, which is an indicator of the competition difficulty and a sign of room for improvement. To achieve top performance, finalist teams used open-source deep learning libraries that were commonly supported with GPUs.


\vspace{5.0mm}

\noindent \textbf{Participants' opinions:} The VIP Cup 2017 created an opportunity for students to work on a real-world problem related to disruptive autonomous vehicle technologies. Even though participants had limited experience, time, and resources, they delivered promising algorithms, had successful presentations, and most importantly, showed unceasing dedication during the competition. Bangladesh University of Engineering and Technology had a finalist team or honorable mention in every SP Cup organized since 2014. Impressively but not surprisingly, two of the finalists are from the Bangladesh University of Engineering and Technology. In this section, we share the inspirational stories and opinions of finalist teams about the 2017 VIP Cup experience.

\vspace{5.0mm}

\noindent \textbf{Team Neurons}

Team \texttt{Neurons} was formed by third year undergraduate students in the Department of Electrical and Electronic Engineering at Bangladesh University of Engineering and Technology. Supervisor of team Neurons professor Md. Kamrul Hasan was also the supervisor of a finalist team in SP Cup, which is a great example of success when mentorship is combined with dedicated students. Experience of team \texttt{Neurons} can best be described by their own words as follows:

\begin{itemize}[label=\textcolor{orange}{\FilledSmallSquare},leftmargin=*]
  
  \item \textit{``It was quite exciting when we heard the news of IEEE VIP Cup being organized this year. We decided to participate as we thought that it would be an excellent opportunity to test our new skill-set and gain some valuable experience doing real-world research... We had quite a lot of hardware limitations...  Nevertheless, we went on with our simulations in spite of the difficulties. Before our Ramadan vacation started, we got access to a PC with a decent CPU and GPU configuration... After that, there was no holding us back. We worked round the clock, often 15-16 hours a day, for 3 whole weeks. A few days before submission, we decided to add another network to our pipeline as some of the test results were not satisfactory. It was a hectic time. Even after submission, the battle wasn’t over. We spent hours on end, in the middle of the night, exchanging emails with our reviewer and providing clarifications for different parts of our submission package. We had to do all of this in the midst of our term-final, handling a lot of academic pressure at the same time. So, it was a great feeling of triumph, when we were selected as one of the finalists and finally, the Champions of the IEEE VIP Cup 2017 - a feat none of us thought we could achieve when we started off in this journey. It was one of the happiest moments of our lives."} - Team \texttt{Neurons} \\
  
  \item ``\textit{The first edition of IEEE VIP CUP was full of 'first ever' experiences for me - first ever participation in any global competition, first ever research project, first ever international conference and so on. Participating in this competition helped me to learn a lot not only about video and image processing, but also about machine learning. The challenge itself was very complex. To process this huge amount of given dataset was also a tremendous challenge for us. But above all, it was definitely a very exciting and rewarding experience!}"- Uday Kamal \\

  \item ``\textit{Participating in the VIP Cup was a very enriching experience. I learned about cutting-edge image processing and machine-learning techniques like neural networks; I learned how to read research papers and write one, give a technical presentation, collaborate with fellow team-members, and, stay focused and motivated even in adverse situations – all of which are extremely valuable for doing any kind of research work. Besides, this is the first time I attended an international conference, where I met researchers and industry representatives leading the field of image-processing. But, most of all, I got the opportunity to represent my country at a global stage, and compete on par with students from around the world. This really gives you a sense of confidence… that, if we are willing to put in the effort, all of us could achieve things we wouldn’t even dare to think of.}" - Sowmitra Das \\
  
  \item ``\textit{It was really a unique experience for me. ICIP 2017 was the first international conference that I’ve joined, and VIP Cup 2017 was the first international competition I’ve participated in my undergraduate life. I’ve also learned a lot of new things throughout this journey.}" - Abid Abrar \\
  
  \item ``\textit{The problem was interesting and quite challenging at least for the students of undergraduate level. I was hesitating in the beginning because of the team members level in the UG program.....limited hardware resources in the lab for machine learning....but the team relieved me within a few weeks. It was enjoyable to see the spirit of the team.}" - Kamrul Hasan, faculty mentor. \\

\end{itemize}


\noindent \textbf{Team PolyUTS}\\
\texttt{PolyUTS} is an international project team across the ocean formed by junior and senior undergraduate students from the Hong Kong Polytechnic University and the University of Technology Sydney. Team members expressed their opinions about the VIP Cup experience as follows:

\begin{itemize}[label=\textcolor{orange}{\FilledSmallSquare},leftmargin=*]

\item ``\textit{The collaboration between us broke the limitation of time and geology. It was through different social platforms that we consistently communicated with each other. Finally, we were truly delighted that our teamwork was such a great success.}" - Team \texttt{PolyUTS} \\

\item 
``\textit{This competition not just taught me a lot of technical related stuff but also let me experience in how these competitions are conducted. That experience is something I will not forget and remember for the rest of my life.}" - Aung Min \\

\item ``\textit{Detecting traffic signs in the given conditions is really a tough task. Although I had learned something about object detection before, I found it was still hard to solve the problem at the beginning. The competition taught me that standing on the shoulder of giants was always a good starting point. By testing and evaluating algorithms proposed in others’ papers, I gradually knew which methods work for this task. Investigating why and how they work finally instructed me what I should do. I think knowing how to overcome such a difficult problem is the most valuable reward for me. I do appreciate the opportunity provided by the match.}" - Jiawei Zhang \\

\item ``\textit{Participating in VIP Cup 2017 was a great experience for me to learn more about machine learning and how it is implemented. Although it was a really challenging task, we still managed to complete a whole system. There were lots of struggles and confusion during the process, but coming over them was also delightful. I'm very happy to spend my summer time on this competition. }" - Weixi Feng \\

\end{itemize}

\noindent \textbf{Team Markovians}\\
Team \texttt{Markovians} was formed by junior undergraduate students from Bangladesh University of Engineering and Technology. Common interest in video and image processing as well as automation formed strong bounds between team members. As electrical engineering students and technology enthusiasts, team members were thrilled to contribute towards the research and development of a traffic sign detection system, which might be able to aid autonomous vehicles. Team \texttt{Markovians} briefly expressed their opinions about their VIP Cup experience as follows:

\begin{itemize}[label=\textcolor{orange}{\FilledSmallSquare},leftmargin=*]

\item ``\textit{Although we had no prior expertise in the field, we did not stop short. We started out from zero and constantly found ourselves at an impasse. However, the support from our supervisor and mentor and the relentless efforts from all the team members was animating. We spent countless hours, group sessions in weekends, through holidays and even before exams for the project. The huge amount of data we needed to process required powerful hardware which wasn’t available to us. However, our perseverance allowed us to proceed with what we had and at the end we came up with a working solution.}" - Team \texttt{Markovians} \\

\end{itemize}

\noindent \textbf{Organizers' opinions:} 
It has been very inspiring to meet the teams in person and hear their stories. Their experience demonstrated that with dedication one can compete at the global level despite hardship, lack of resources, and limited support. In addition, this experience showed the importance of mentorship exerted here by professors who lead these undergraduate teams to excel. The dedication of competing teams motivated the organizers to work tirelessly throughout the competition.

\vspace{5.0mm}

\noindent \textbf{Upcoming VIP Cup:} The second edition of the VIP Cup will be held at the 2018 \textit{IEEE International Conference on Image Processing} in Athens, Greece, between October 7-10, 2018. The theme of the 2018 competition will be announced in February. Teams who are interested in the VIP Cup can visit: 
\href{https://signalprocessingsociety.org/get-involved/video-image-processing-cup}{https://signalprocessingsociety.org/get-involved/video-image-processing-cup}.

In addition to the VIP Cup, the IEEE Signal Processing Society  announced the fifth edition of the SP Cup. The final competition will be held at the IEEE International Conference on Acoustics, Speech and Singal Processing 2018 in Calgary, Alberta, Canada, 15-20 April 2018. The theme of the competition is ``Forensic Camera Model Identification". For details, visit: \href{https://signalprocessingsociety.org/get-involved/signal-processing-cup}{https://signalprocessingsociety.org/get-involved/signal-processing-cup}.

\noindent \textbf{Acknowledgements:} The organizers of the VIP Cup 2017 would like to acknowledge and express their gratitude to everyone involved in the first VIP Cup journey including but not limited to Patrizio Campisi for initiating the VIP Cup and his unmatched support throughout the competition; Min-Hung Chen and Tariq Alshawi for their significant roles in the organization including dataset preparation and algorithm validation; both Amy Reibman and B\'eatrice Pesquet-Popescu for  serving on the jury; and participating teams for their hard work, dedication, and inspirational stories. Finally, the sponsorship by the MMSP Technical Committee of this Cup is much appreciated.  

\noindent \textbf{Authors} 

\textbf{Dogancan Temel} (cantemel@gatech.com) is a postdoctoral fellow at the Georgia Institute of Technology. He was the recipient of the Best  Doctoral  Dissertation  Award from Sigma Xi honor society, the Graduate Research Assistant Excellent Award from the School of ECE, and the Outstanding Research Award from the Center for Signal and Information Processing at Georgia Tech. \\   

\textbf{Ghassan AlRegib} (alregib@gatech.edu) is  currently  a  Professor  in  the School   of   Electrical   and   Computer   Engineering at the Georgia Institute of Technology. He  is  the  director  of Omni Lab for Intelligent Visual Engineering and Science (OLIVES) and the Center  for  Energy and Geo  Processing (CeGP). He is  a  Member  of  the  IEEE  SPS  MMSP and IVMSP Technical  Committees.  He served at various capacities within the IEEE SPS Society including the Technical Program co-Chair of GlobalSIP 2014, Tutorial  Chair  for  ICIP16, and Technical Program Chair of ICIP 2020. He received various awards including the Outstanding Junior Faculty Award in 2008 and the Global Engagement Excellence Award in 2017. \\




\ifCLASSOPTIONcaptionsoff
  \newpage
\fi






\end{document}